\documentclass{article}

\usepackage{microtype}
\usepackage{graphicx}
\usepackage{subcaption}
\usepackage{booktabs} 

\usepackage{hyperref}

\usepackage[preprint]{icml2026}

\usepackage{amsmath}
\usepackage{amssymb}
\usepackage{mathtools}
\usepackage{amsthm}

\usepackage[capitalize,noabbrev]{cleveref}

\theoremstyle{plain}

\theoremstyle{definition}

\theoremstyle{remark}

\usepackage[textsize=tiny]{todonotes}

\usepackage{multirow, makecell}
\usepackage{enumitem}
\usepackage{xspace}
\usepackage{threeparttable}
\usepackage{titlesec}

\makeatletter
\newcommand*{\eg}{e.g.\@\xspace}

\makeatother

\icmltitlerunning{Beyond Fixed Frames: Dynamic Character-Aligned Speech Tokenization}

\begin{document}
\setlength{\abovedisplayskip}{3pt}
\setlength{\belowdisplayskip}{3pt}
\setlength{\abovedisplayshortskip}{2pt}
\setlength{\belowdisplayshortskip}{2pt}

\setlength{\parskip}{1pt}
\setlength{\parindent}{0pt}


\twocolumn[
  \icmltitle{Beyond Fixed Frames: Dynamic Character-Aligned Speech Tokenization}



  \icmlsetsymbol{equal}{*}

  \begin{icmlauthorlist}
    \icmlauthor{Luca {Della Libera}}{concordia,mila}
    \icmlauthor{Cem Subakan}{laval,concordia,mila}
    \icmlauthor{Mirco Ravanelli}{concordia,mila}
  \end{icmlauthorlist}

    \icmlaffiliation{concordia}{Concordia University, Montreal, Canada}
    \icmlaffiliation{mila}{Mila-Quebec AI Institute, Montreal, Canada}
    \icmlaffiliation{laval}{Université Laval, Québec, Canada}
    
    \icmlcorrespondingauthor{Luca {Della Libera}}{luca.dellalibera@mail.concordia.ca}

\icmlkeywords{Speech coding, Variable frame rate, Vector quantization, Generative models}
  \vskip 0.3in
]


\printAffiliationsAndNotice{}  

\begin{abstract}
\vspace{3.0pt}
Neural audio codecs are at the core of modern conversational speech technologies, converting continuous speech into sequences of discrete tokens that can be processed by LLMs. However, existing codecs typically operate at fixed frame rates, allocating tokens uniformly in time and producing unnecessarily long sequences. In this work, we introduce \textbf{DyCAST}, a \textbf{Dy}namic \textbf{C}haracter-\textbf{A}ligned \textbf{S}peech \textbf{T}okenizer that enables variable-frame-rate tokenization through soft character-level alignment and explicit duration modeling. DyCAST learns to associate tokens with character-level linguistic units during training and supports alignment-free inference with direct control over token durations at decoding time. To improve speech resynthesis quality at low frame rates, we further introduce a retrieval-augmented decoding mechanism that enhances reconstruction fidelity without increasing bitrate. Experiments show that DyCAST achieves competitive speech resynthesis quality and downstream performance while using significantly fewer tokens than fixed-frame-rate codecs.
Code and checkpoints will be released publicly at \href{https://github.com/lucadellalib/dycast}{https://github.com/lucadellalib/dycast}.
\end{abstract}

\section{Introduction}
In the past few years, tokenization has emerged as a key building block in multimodal LLMs, enabling joint autoregressive modeling across modalities~\cite{dubey2024llama3herdmodels,jiang2024mixtralexperts,comanici2025gemini25,singh2025openaigpt5card,deepseekai2025deepseekv3}. By representing diverse modalities as sequences of discrete tokens, these models can operate over text, images, and audio within a shared modeling framework.
For speech, tokenization is typically achieved through \textbf{neural audio codecs}, which map continuous waveforms to discrete tokens. Such representations have driven recent progress in generative speech and audio modeling~\cite{borsos2023audiolm, wang2023valle, nguyen2024spiritlminterleavedspokenwritten, defossez2024moshi, kyutai2025hibiki, zeghidour2025delayed}, enabling LLMs to operate directly on speech. Originally developed for efficient transmission~\cite{zeghidour2021soundstream,defossez2023encodec}, modern codecs now produce compact yet powerful representations that support high-quality resynthesis as well as competitive performance in both discriminative and generative downstream tasks~\cite{mousavi2025dates, guo2025discretespeechtokensreview}.

Despite their success, most existing speech tokenizers rely on frame-level representations at a \textbf{fixed frame rate}, producing tokens at a constant temporal resolution. While effective for reconstruction, this design is poorly matched to the inherently variable temporal structure of speech: silence and steady regions are information-poor, whereas rapidly changing segments are information-dense~\cite{vankuyk2017information,dieleman2021variable,cuervo2022cpc}. As a result, fixed-frame-rate tokenization leads to inefficient sequence lengths and limited alignment with corresponding text, making generative modeling more challenging.
Although \textbf{dynamic-frame-rate} codecs and \textbf{text-supervised alignment} methods have recently emerged, this area is still developing. Existing approaches often rely on heuristic frame merging or clustering strategies that are weakly grounded in linguistic structure~\cite{wang2025codecslime, zhang2025unlocking, zheng2025say, li2025flexicodec}, or require transcriptions or alignment information at inference time~\cite{tseng2025tastetextalignedspeechtokenization, hsu2025taslatextalignedspeechtokens}, limiting flexibility in fully speech-based scenarios.

\begin{figure*}[t!]
  \centering
\includegraphics[width=0.77\textwidth]{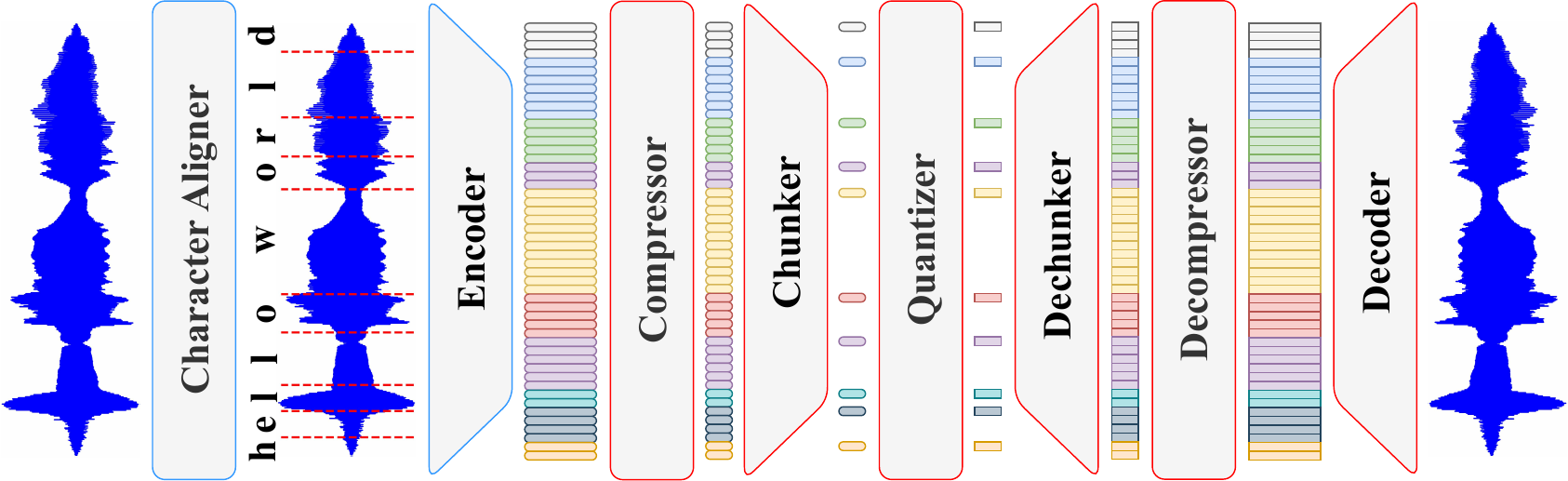}
  \vspace{-0.1cm}
  \caption{DyCAST architecture. Frame-level representations extracted by a frozen, self-supervised encoder are compressed and dynamically grouped into variable-length chunks, pooled, and quantized into discrete tokens. The decoding stage reverses this process by expanding token-level representations back to frame-level features before waveform reconstruction. Character-level boundaries are provided during training by a frozen aligner.}
  \label{fig:dycast}
\vskip -0.2in
\end{figure*}

In this work, we introduce \textbf{DyCAST}, a \textbf{Dy}namic \textbf{C}haracter-\textbf{A}ligned \textbf{S}peech \textbf{T}okenization framework that enables variable-frame-rate coding through \textbf{soft character-level alignment}.
Each speech token is approximately associated with a character in the underlying transcript, with the degree of alignment controlled by a learned \textbf{boundary predictor}. At inference time, this alignment can be adjusted dynamically, allowing a flexible trade-off between strict character alignment and longer token spans.
As a result, DyCAST can operate either with ground-truth alignments when available (\eg for text-to-speech) or in a fully alignment-free mode based on the learned soft alignment alone. Additionally, DyCAST provides explicit control over token durations during decoding via a \textbf{duration predictor} that learns to invert the pooling operation, predicting appropriate token durations from the context.
Because character-level representations naturally result in very low frame rates, achieving high-quality resynthesis becomes more challenging. To address this limitation, we further introduce \textbf{retrieval-augmented decoding} as an auxiliary mechanism to improve reconstruction quality by leveraging side information, without increasing bitrate.

In summary, our contributions are as follows:
\begin{itemize}[leftmargin=*, topsep=0pt, itemsep=0pt, parsep=0pt, labelsep=0.3em, labelindent=0pt, itemindent=0pt]
\item We propose {DyCAST}, a novel framework for \textbf{controllable}, \textbf{character-aligned}, \textbf{variable-frame-rate} speech tokenization through soft character-level alignment and explicit duration modeling.
\item We introduce \textbf{retrieval-augmented decoding} as an auxiliary mechanism to enhance speech resynthesis quality by leveraging side information at inference time.
\item We empirically show \textbf{competitive performance} on speech resynthesis and a range of other downstream tasks, while using significantly \textbf{fewer tokens} compared to fixed-frame-rate baselines.
\end{itemize}

Code and checkpoints will be released publicly at \href{https://github.com/lucadellalib/dycast}{https://github.com/lucadellalib/dycast}.


\section{Related Work}

\paragraph{Fixed-Frame-Rate Codecs.}
The dominant approach to speech tokenization discretizes audio at a fixed temporal resolution using frame-level tokens. Early research on neural speech codecs primarily focused on high-fidelity {acoustic} reconstruction at medium bitrates~\cite{zeghidour2021soundstream, defossez2023encodec, kumar2023dac}. In parallel, a separate line of work explored {semantic} tokenization by discretizing self-supervised phonetic speech representations to capture linguistic content~\cite{baevski2020wav2vec2, hsu2021hubert, messica24nast, mousavi2024how, chang2024dcspins}, at the expense of fine-grained acoustic detail.
More recently, {hybrid} codecs have emerged with the goal of balancing semantic and acoustic information. These approaches combine multiple design strategies, including the use of multiple codebooks~\cite{ju2024facodec, jiang2024unicodec, zheng2024freecodec}, dual-encoder architectures~\cite{liu2024semanticodec}, knowledge distillation~\cite{zhang2024speechtokenizer, defossez2024moshi, yang2025almtokenizer, gong2025xytokenizermitigatingsemanticacousticconflict, li2025dualcodec}, or supervised fine-tuning~\cite{hartuv2025past}, to improve both reconstruction quality and speech language modeling performance.
More recently, a growing trend has shifted toward {single-codebook} designs that jointly encode semantic and acoustic information. These approaches achieve good reconstruction quality at low bitrates while substantially simplifying downstream modeling~\cite{ji2024wavtokenizer, bai2024dmel, xin2024bigcodec, wu2024ts3codec, ye2025llasa, dellalibera2025focalcodec, dellalibera2025focalcodecstream, song2025magicodec}.
Our work departs from fixed-rate tokenization by introducing dynamic, character-aligned speech tokens whose durations adapt to linguistic content, providing more efficient and linguistically grounded representations.

\paragraph{Variable-Frame-Rate Codecs.}
Dynamically adjusting frame rates based on content complexity has recently attracted attention across modalities. In the text domain, several recent works~\cite{nawrot2023dtp,slagle2024spacebytedeletingtokenizationlarge, ahia2024magnet,pagnoni2025blt,videau2025bytesideaslanguagemodeling,hwang2025dynamicchunkingendtoendhierarchical} explore {learned tokenization} approaches, in which chunking is learned end-to-end according to a task objective, rather than being fixed a priori. In these methods, token boundaries emerge from the optimization process itself, enabling adaptive representations that better match the structure of the underlying data.
A similar trend is emerging in audio.
Early work by \citet{dieleman2021variable} proposed an audio VQ-VAE combined with run-length encoding to enable variable frame rates. More recently, CodecSlime~\cite{wang2025codecslime} introduced a multi-stage pipeline in which a fixed-frame-rate codec is first trained and then temporally similar representations are merged to obtain adaptive frame rates. TFC~\cite{zhang2025unlocking} and VARSTok~\cite{zheng2025say} propose dynamic-frame-rate strategies in which duration is encoded implicitly through the structure of the codebook. FlexiCodec~\cite{li2025flexicodec} further builds on VARSTok by enabling substantially lower frame rates through refined merging procedures, while retaining fine-grained control over the resulting token rate.
Unlike prior approaches that rely on heuristic frame merging or require transmitting durations, our method uses learned character-level soft alignment to produce semantically aligned variable-rate tokens without duration side information, while still enabling explicit duration control at decoding time.


\paragraph{Text-Aligned Speech Representations.}
Recent work has explored text-aligned tokenization as a way to bridge the modality gap for joint speech-text modeling. Methods such as TASTE~\cite{tseng2025tastetextalignedspeechtokenization} and TASLA~\cite{hsu2025taslatextalignedspeechtokens} learn discrete speech tokens aligned with text by enforcing transcript-speech alignment through cross-attention-based architectures.
Similarly, TaDiCodec~\cite{wang2025tadicodec} incorporates textual information in a text-aware diffusion-based speech codec, where speech-text alignment is introduced implicitly through autoregressive conditioning on text rather than being enforced explicitly at the tokenization stage.
SSR~\cite{tan2025ssr} and LST~\cite{lu2025latentspeechtexttransformer} pursue a similar goal of speech-text alignment, but are tangential to our work, as they operate at the speech language modeling level and do not define a standalone speech tokenization framework.
While effective for joint modeling, these methods typically rely on ground-truth alignments or require textual input at inference time, limiting their applicability to text-free generative settings and making them closer to TTS-oriented pipelines than general-purpose speech tokenizers. In contrast, DyCAST is a fully non-autoregressive, autoencoder-based speech tokenizer that operates end-t-o end without requiring text at inference time.

\section{DyCAST}
The proposed DyCAST framework (see \cref{fig:dycast}) is inspired by the modular compressor-quantizer-decompressor architecture introduced in~\cite{dellalibera2025focalcodec}, while extending it with dedicated modules for \emph{dynamic pooling}.
Given an input waveform, a frozen, pretrained self-supervised speech \textbf{encoder} first extracts high-dimensional, single-stream acoustic-semantic representations at a fixed frame rate. These features are then projected into a lower-dimensional latent space by a lightweight \textbf{compressor}, retaining the most informative components of the representation.
A \textbf{chunker} module subsequently groups consecutive frames into variable-length chunks, yielding a temporally adaptive chunking of the input. Within each chunk, frame-level features are pooled to form compact chunk-level representations, which are then discretized by a \textbf{quantizer}, producing a sequence of discrete tokens. A \textbf{dechunker} module reverses the chunking operation by expanding each token-level representation back to frame-level features. A \textbf{decompressor} maps the expanded features back to the original encoder dimensionality, and a \textbf{decoder} finally reconstructs the time-domain waveform.

\subsection{Dynamic Chunking}
\label{subsec:chunker}
The \emph{chunker} module consists of two components: a \textbf{boundary predictor}, which identifies semantically meaningful chunk boundaries, and a \textbf{downsampler}, which aggregates frame-level representations within each chunk into a compact chunk-level representation.
To provide training supervision for chunk boundaries, we employ a frozen, pretrained \textbf{character aligner}. This aligner takes a waveform as input and returns character-level durations, and can be conveniently implemented using a CTC-based ASR model~\cite{graves2006connectionist} with characters as the vocabulary. The resulting frame-level character boundaries are used to derive target chunk durations that serve as supervision during training. To model the boundary distribution, we adopt a \textbf{discrete-time hazard model}~\cite{singer1993hazard, ren2019survival} rather than a standard frame-wise binary cross-entropy objective. Unlike independent boundary classification, the hazard formulation explicitly models the \emph{time to the next boundary}, allowing boundary predictions to be temporally dependent and properly normalized over time. This is particularly important in our setting, where chunk durations are the primary quantity of interest and boundaries are sparse relative to the frame rate. Moreover, the hazard model naturally enforces a single boundary per chunk and provides a principled likelihood over variable-length segments, which facilitates stable training and coherent boundary decoding.
Formally, given frame-level representations $x_{1:T}$ extracted by the compressor, the boundary predictor estimates a boundary probability $h_t \in (0,1)$ at each frame:
\begin{equation}
h_t = \sigma\!\left(f_\theta(x_{1:T})_t\right),
\end{equation}
where $f_\theta$ is neural network operating on the sequence of features. The probability that the next boundary occurs $k$ frames after time $t$ is then given by
\begin{equation}
P(T = k \mid t) = \left( \prod_{i=0}^{k-1} (1 - h_{t+i}) \right) h_{t+k}.
\end{equation}
The hazard model is trained by maximizing the likelihood of ground-truth next-boundary offsets extracted from forced alignment.

At inference time, chunk boundaries are decoded directly from the hazard predictions without any textual input. Boundaries can be obtained either greedily, yielding deterministic chunking, or via sequential sampling to introduce variability. In both cases, decoding constraints on the minimum and maximum chunk duration, controlled by the hyperparameters \texttt{min\_gap} and \texttt{max\_gap}, respectively, are used to regulate the resulting frame rate. In addition, a hazard threshold $\tau_h$ controls boundary emission: higher values of $\tau_h$ favor longer chunks (lower frame rates), while lower values produce finer-grained chunks (higher frame rates). Formally, a boundary is emitted at frame $t$ when $h_t \ge \tau_h$ subject to the \texttt{min\_gap} constraint; if no boundary is emitted for \texttt{max\_gap} consecutive frames, a boundary is forced.
Given the decoded boundaries, the downsampler produces chunk-level representations by selecting the \emph{last frame} of each chunk. While alternative strategies such as average pooling per chunk are possible, this design keeps the operation simple and efficient, and preserves the original compressed representations without blending them across frames, which facilitates adaptation to different frame rates.

\subsection{Dechunking}
\label{subsec:dechunking}
Like the chunker, the \emph{dechunker} module consists of two components: a \textbf{duration predictor}, which estimates the number of frames associated with each character-aligned discrete token, and an \textbf{upsampler}, which expands each chunk-level representation back to frame-level features.
The duration predictor is necessary to recover temporal structure at decoding time, since in general only the token sequence is transmitted, not the chunk boundaries used to derive it.
Supervision for duration prediction comes from the same frozen character aligner used during chunking, which provides target token durations corresponding to character spans.
To model the distribution of token durations, we adopt a \textbf{negative binomial duration model}~\cite{zen2009statistical}. Compared to alternatives such as geometric or Poisson distributions, the negative binomial provides greater flexibility by explicitly modeling over-dispersed count data, which is characteristic of speech durations. In particular, while a geometric model assumes a memoryless process with a fixed hazard rate and a Poisson model ties the mean and variance, the negative binomial decouples these quantities, allowing the model to capture the heavy-tailed and highly variable nature of character-level durations in speech.
Formally, given a sequence of discrete tokens $c_{1:N}$, the duration model maps each token to a positive \emph{free} mean duration $\mu_i^{\mathrm{free}} > 0$,
\vspace{-0.25cm}
\begin{equation}
\mu_i^{\mathrm{free}} = \mathrm{softplus}\!\left(g_\phi(c_{1:N})_i\right),
\end{equation}
where $g_\phi$ denotes a neural network operating on the full sequence of tokens.
The actual mean token duration is obtained by enforcing a minimum duration $d_{\min}$,
\begin{equation}
\mu_i = d_{\min} + \mu_i^{\mathrm{free}}.
\end{equation}
By default, $d_{\min} = 1$. The excess target duration
\vspace{-0.05cm}
\begin{equation}
y_i = d_i - d_{\min} \ge 0
\end{equation}
is modeled using the negative binomial distribution with mean $\mu_i^{\mathrm{free}}$ and a global dispersion parameter $\alpha > 0$ shared across all tokens,
\vspace{-0.25cm}
\begin{equation}
y_i \sim \mathrm{NB}(\mu_i^{\mathrm{free}}, \alpha).
\end{equation}

The duration model is trained by minimizing the negative log-likelihood together with a \textbf{normalized length regularization} term,
\vspace{-0.1cm}
\begin{equation}
\fontsize{8.5}{9.5}
\mathcal{L}_{\mathrm{dur}} =
\sum_{i=1}^{N}
-\log p_{\mathrm{NB}}(y_i \mid \mu_i^{\mathrm{free}}, \alpha)
\;+\;
\lambda
\left(
\frac{\sum_{i=1}^{N} \mu_i^{\mathrm{free}} - T_{\mathrm{free}}}{T_{\mathrm{free}} + \epsilon}
\right)^2,
\end{equation}
where $T_{\mathrm{free}} = T - N d_{\min}$ denotes the total number of allocatable frames after enforcing a minimum duration $d_{\min}$ per token, $\lambda$ controls the strength of the length regularization, and $\epsilon$ is a small constant for numerical stability. This normalized penalty encourages globally consistent pacing while remaining invariant to utterance length.

At inference time, the duration model supports two decoding regimes. In the \textbf{free decoding} mode, when the total number of frames is unknown, durations are obtained greedily as
\begin{equation}
\hat d_i = d_{\min} + \mathrm{round}(\mu_i^{\mathrm{free}}),
\end{equation}
yielding a predicted total length $\hat T = \sum_i \hat d_i$. In the \textbf{budget-constrained decoding} mode, when the target length $T$ is known (\eg speech resynthesis), the predicted free means are renormalized to match the available duration budget,
\begin{equation}
\tilde{\mu}_i^{\mathrm{free}} =
\mu_i^{\mathrm{free}}
\frac{T_{\mathrm{free}}}{\sum_j \mu_j^{\mathrm{free}}},
\end{equation}
followed by deterministic integer rounding that enforces $\sum_i \hat d_i = T$ exactly. This hybrid formulation enables explicit duration control at decoding time while remaining independent of character alignment during inference.

\subsection{Retrieval-Augmented Decoding}
\begin{figure}[t!]
  \centering
\includegraphics[width=0.36\textwidth]{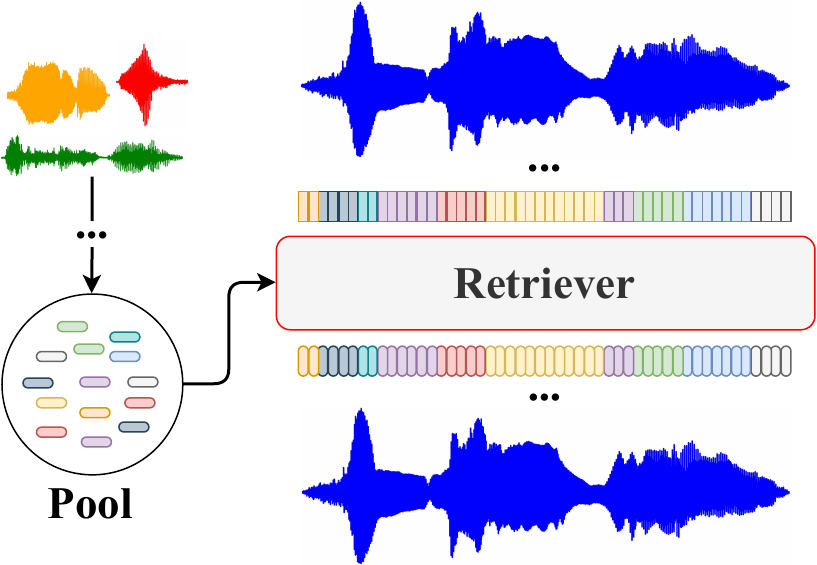}
  \vspace{-0.1cm}
  \caption{Retrieval-augmented decoding (RAD). Discrete latents are refined via similarity search against a pool of continuous latents prior to waveform reconstruction.}
  \label{fig:rad}
\vskip -0.27in
\end{figure}
Since the proposed codec is character-aligned, the resulting frame rate is naturally very low (6 -- 18 Hz). This makes accurate waveform reconstruction increasingly challenging, as fine-grained acoustic details, high-frequency information, and speaker-specific characteristics tend to be removed by the quantization bottleneck~\cite{yang2025almtokenizer, li2025flexicodec}.
To mitigate this issue without increasing the bitrate, we propose \textbf{retrieval-augmented decoding} (RAD), a decoder-side mechanism that improves reconstruction quality by selectively incorporating auxiliary information at decoding time (see \cref{fig:rad}).
RAD builds on the fact that self-supervised speech encoders trained with masking objectives~\cite{baevski2020wav2vec2,hsu2021hubert,chen2022wavlm} produce intermediate representations with strong semantic structure. In these models, representations that are nearby in feature space tend to correspond to similar linguistic content, enabling meaningful retrieval at the latent level via similarity search. This property underlies the effectiveness of nearest-neighbor methods in tasks such as voice conversion~\cite{baas2023knnvc,dellalibera2025focalcodec}.
RAD leverages this structure by maintaining a \textbf{pool of continuous latents} collected from a diverse set of utterances, speakers, and acoustic conditions. During decoding, this pool can be queried to retrieve similar latent vectors. By analogy to retrieval-augmented generation, where a query embedding retrieves relevant documents from a latent knowledge store, RAD retrieves continuous speech latents from this pool via similarity matching against the discrete latent produced by the quantizer. Since these latents are already low-dimensional and compact by design, they are well-suited for efficient nearest-neighbor search and large-scale retrieval.
For each quantized latent, if the similarity to its nearest retrieved candidate exceeds a predefined threshold $\tau$, the discrete latent is replaced with the retrieved continuous latent. By setting a sufficiently high similarity threshold, retrieval can be restricted to candidates that preserve \textbf{both semantic content and speaker identity}. In cases where near-identical speech segments exist in the pool, RAD can recover the original continuous representation, substantially improving speaker fidelity and fine-grained acoustic detail.
Interestingly, the candidate pool can be constructed offline and updated over time to adapt to evolving deployment needs (\eg privacy, memory footprint, speed, and/or efficiency).

\section{Experimental Setup}

\subsection{Architecture}
We now describe the DyCAST architecture used in our experiments, inspired by FocalCodec~\cite{dellalibera2025focalcodec}. Full hyperparameter specifications are reported in Appendix~\ref{sec:hyperparameters}.

\paragraph{Character Aligner.}
We employ MMS~\cite{pratap2024mms}. Alignment information is extracted via simple \texttt{argmax} CTC alignment on the frame-level log probabilities obtained by running the model on the input waveform.
To handle silence segments detected by the aligner, we do not discard them; instead, we aggregate them into the subsequent non-silence token. We find this strategy to work well in practice and leave more refined treatments of silence for future work.

\paragraph{Encoder and Decoder.}
We use the pretrained WavLM-large\footnote{\href{https://github.com/microsoft/unilm/tree/master/wavlm}{https://github.com/microsoft/unilm/tree/master/wavlm}} model~\cite{chen2022wavlm} as the encoder. In particular, we use the representations from the \textbf{6th transformer layer}, which are semantically rich while retaining fine-grained acoustic information~\cite{dellalibera2025focalcodec, baas2023knnvc}.
Waveform reconstruction from these features is performed using Vocos~\cite{siuzdak2023vocos} for efficient decoding.

\paragraph{Compressor and Decompressor.}
Both modules are implemented using focal modulation blocks~\cite{yang2022focalnets,dellaliber2024focal}, leveraging their demonstrated effectiveness and efficiency for low-bitrate speech coding~\cite{dellalibera2025focalcodec}. 

\paragraph{Boundary and Duration Predictor.}
We adopt the same efficient focal modulation architecture also for the boundary and duration predictor. The boundary predictor uses a binary classification head, while the duration model predicts a free mean parameter together with a shared, learnable dispersion parameter $\alpha$, as described in \cref{subsec:dechunking}. Importantly, the boundary predictor is trained directly on the 1024-dimensional WavLM features rather than the compressed latents, providing richer context for boundary estimation. In contrast, the duration predictor operates on the pooled, quantized low-dimensional latents.
\vspace{-0.15cm}

\paragraph{Quantizer.}
\citet{dellalibera2025focalcodec} employ binary spherical quantization (BSQ)~\cite{zhao2024bsq}, which tightly couples bitrate to latent dimensionality $L$ through an implicit codebook of size $|\mathcal{C}| = 2^L$. At the low frame rates considered in this work, this coupling limits representational expressivity.
We therefore generalize this formulation to \textbf{scalar spherical quantization} (SSQ). SSQ preserves the spherical constraint of BSQ while allowing each latent dimension to take one of $K$ scalar levels in $[-\frac{1}{\sqrt{L}}, \frac{1}{\sqrt{L}}]$, enabling more flexible bitrates. Compared to finite scalar quantization (FSQ)~\cite{mentzer2024finite}, SSQ explicitly enforces spherical geometry via $L_2$ normalization and includes a factorized entropy regularizer for improved codebook utilization. The resulting implicit codebook has size $|\mathcal{C}| = K^L$. In our experiments, we set $L=32$ and $K=4$, yielding $|\mathcal{C}| = 4^{32}$. To make this high cardinality tractable in practice, we adopt a factorized representation in which each token is decomposed into 32 parallel streams, each with a vocabulary size of 4.\looseness=-1

\subsection{Training}
We train DyCAST on LibriTTS~\cite{zen2019libritts}, resampled to 16\,kHz. We adopt a multi-stage training strategy for improved stability:
\begin{enumerate}[leftmargin=*, topsep=0pt, itemsep=0pt, parsep=0pt, labelsep=0.3em, labelindent=0pt, itemindent=0pt]
    \item \textbf{Reconstruction.} We train the compressor-quantization-decompressor for WavLM features reconstruction using $L_2$ loss, \emph{without} boundary prediction, or duration modeling. Dynamic downsampling and upsampling use the character durations from the character aligner in a teacher-forced manner. In parallel, we train the decoder to map WavLM features back to waveform.
    \item \textbf{Boundary predictor training.} We train the boundary predictor to map WavLM features to character boundaries, using character aligner durations as supervision.
    \item \textbf{Adaptation to predicted boundaries.} We fine-tune the compressor-quantizer-decompressor to operate on durations predicted by the boundary predictor. To improve robustness, during training we randomly sample among character aligner boundaries and predicted boundaries greedily decoded with $\tau_h = 0.5$, \texttt{min\_gap} $\in \{1,3,5\}$, and no \texttt{max\_gap} constraint (see \cref{subsec:chunker}).
    \item \textbf{Duration predictor training.} Finally, we freeze all the other components and train the duration model to predict chunk durations from pooled quantized latents, using the same boundary sampling strategy as in the previous step. This procedure yields a robust system that generalizes well to different boundary configurations, including \texttt{min\_gap} values outside the training range.
\end{enumerate}
Hyperparameters and training details are provided in Appendix~\ref{sec:hyperparameters}.

\subsection{Inference}
DyCAST provides substantial flexibility at inference time for both encoding and decoding.
For encoding, chunk boundaries can be obtained in two ways:
\begin{enumerate}[leftmargin=*, topsep=0pt, itemsep=0pt, parsep=0pt, labelsep=0.3em, labelindent=0pt, itemindent=0pt]
\item \textbf{Character aligner.} We use boundaries provided by the character aligner. This is particularly useful for downstream applications that benefit from accurate speech-text alignment, such as text-to-speech.
\item \textbf{Boundary predictor.} We greedily decode or sample boundaries from the boundary predictor. By adjusting the decoding hyperparameters, we obtain direct control over the effective frame rate. For example, increasing the \texttt{min\_gap} parameter enforces a larger minimum gap between boundaries, resulting in fewer, more uniformly spaced tokens.
\end{enumerate}

For decoding, three operating modes are available:
\begin{enumerate}[leftmargin=*, topsep=0pt, itemsep=0pt, parsep=0pt, labelsep=0.3em, labelindent=0pt, itemindent=0pt]
    \item \textbf{Tokens + durations.} We transmit discrete tokens together with their original durations and use them as side information during decoding. The effective bitrate is the token bitrate plus the overhead required to transmit one duration per token, assuming a bounded duration range.
    \item \textbf{Tokens + utterance length.} We transmit discrete tokens and a single global utterance length, which incurs negligible overhead. The duration model is then used in budget-constrained decoding mode to sample durations whose total length matches the provided value. This mode is particularly suitable for resynthesis tasks.
    \item \textbf{Tokens only.} We transmit only discrete tokens and let the duration model infer the most likely durations. This mode is useful for modeling tasks such as text-to-speech and speech language modeling, where tokens do not come with explicit duration information.
\end{enumerate}
Note that, independently of the chosen setting, durations can always be resampled using the duration model or an external model to introduce natural variability in speaking rate, or manually overridden to stretch or compress individual characters. Overall, this design offers significantly greater flexibility than traditional fixed-rate codecs, enabling fine-grained control over bitrate, timing, and prosody.

\section{Downstream Evaluation}
For downstream evaluation, we consider four inference configurations: \textbf{DyCAST-CA}, \textbf{DyCAST-BP1}, \textbf{DyCAST-BP3}, and \textbf{DyCAST-BP5}. DyCAST-CA uses boundaries provided by the character aligner, while the remaining variants rely on the boundary predictor with \texttt{min\_gap} set to 1, 3, and 5, respectively, corresponding to average frame rates of approximately 14 Hz, 9 Hz, and 6 Hz.
Unless otherwise stated, all experiments use the \emph{tokens + utterance length} decoding mode.
We provide analysis of alternative decoding strategies and visualizations in Appendix~\ref{sec:appendix_decode_mode} and \ref{sec:appendix_spectrograms}.

To showcase the effectiveness of our framework, we compare against a range of fixed-frame-rate baselines (see \cref{tab:baselines}).
Additional details about these baselines are provided in Appendix~\ref{sec:appendix_baselines_codecs}.
We follow the evaluation protocol of \citet{dellalibera2025focalcodec}, considering speech resynthesis under different conditions, voice conversion, and both discriminative and generative downstream tasks. For generative evaluation, we focus on text-to-speech in limited-data regimes.
Details about the datasets and experimental settings can be found in Appendix~\ref{sec:appendix_datasets} and \ref{subsec:appendix_downstream}.


\begin{table}[t!]
\vspace{-0.1cm}
\setlength{\tabcolsep}{2pt}
\caption{Codecs considered in our downstream evaluation.}
\label{tab:baselines}
\vspace{-0.35cm}
\begin{center}
\begin{footnotesize}
\resizebox{0.37\textwidth}{!}{%
\begin{tabular}{lcccccc}
\toprule
\textbf{Codec} &
\makecell{\textbf{Frame} \\ \textbf{Rate} \textbf{(Hz)}} &
\makecell{\textbf{Bitrate} \\ \textbf{(kbps)}} &
\makecell{\textbf{Sample} \\ \textbf{Rate} \textbf{(kHz)}} &
\makecell{\textbf{Codebooks}} &
\makecell{\textbf{Params} \\ \textbf{(M)}} \\
\midrule
EnCodec & 75.0 & 1.50 & 24 & $2 \times 1024$ & 15 \\
DAC & 50.0 & 1.00 & 16 & $2 \times 1024$ & 74 \\
WavLM6-KM & 50.0 & 0.45 & 16 & $1 \times 512$ & 127 \\
SpeechTokenizer & 50.0 & 1.00 & 16 & $2 \times 1024$ & 108 \\
SemantiCodec & 25.0 & 0.65 & 16 & $2 \times 8192$ & 1033 \\
Mimi & 12.5 & 0.69 & 24 & $5 \times 2048$ & 82 \\
WavTokenizer & 40.0 & 0.48 & 24 & $1 \times 4096$ & 85 \\
BigCodec & 80.0 & 1.04 & 16 & $1 \times 8192$ & 160 \\
Stable Codec & 25.0 & 0.70 & 16 & $2 \times 15625$ & 950 \\
FocalCodec & 50.0 & 0.65 & 16 & $1 \times 8192$ & 142 \\
\midrule
\textbf{DyCAST-CA} & $\sim$14.4 & $\sim$0.92
& 16 & $1 \times 4^{32}$ & 269 \\

\textbf{DyCAST-BP1} & $\sim$17.5 & $\sim$1.12
& 16 & $1 \times 4^{32}$ & 269 \\

\textbf{DyCAST-BP3} & $\sim$9.0 & $\sim$0.57 
& 16 & $1 \times 4^{32}$ & 269 \\

\textbf{DyCAST-BP5} & $\sim$6.2 & $\sim$0.40
& 16 & $1 \times 4^{32}$ & 269 \\
\bottomrule
\end{tabular}
}
\end{footnotesize}
\end{center}
\vskip -0.28in
\end{table}

\begin{table*}[t!]
\setlength{\tabcolsep}{2pt}
\caption{Speech resynthesis (SR) and voice conversion (VC). \textbf{Best} and \underline{second-best} results are highlighted.}
\vskip -0.15in
\label{tab:speech_resynthesis}
\begin{center}
\begin{footnotesize}
\resizebox{0.935\textwidth}{!}{%
\begin{tabular}{lc c|ccc|ccc|ccc|ccc|ccc}
\toprule
\multirow{3}{*}{\textbf{Codec}} &
\multirow{3}{*}{\makecell{\textbf{Frame} \\ \textbf{Rate}  \textbf{(Hz)}} $\downarrow$} &
\multirow{3}{*}{\makecell{\textbf{Bitrate} \\ \textbf{(kbps)}} $\downarrow$}
& \multicolumn{3}{c|}{\textbf{SR -- LibriSpeech}}
& \multicolumn{3}{c|}{\textbf{SR -- MLS}}
& \multicolumn{3}{c|}{\textbf{SR -- VoiceBank}}
& \multicolumn{3}{c|}{\textbf{SR -- Libri1Mix}}
& \multicolumn{3}{c}{\textbf{VC -- VCTK}} \\
\cmidrule{4-18}
&
&
& \textbf{UTMOS}  $\uparrow$ & \textbf{dWER}  $\downarrow$ & \textbf{Sim}  $\uparrow$
& \textbf{UTMOS}  $\uparrow$ & \textbf{dWER}  $\downarrow$ & \textbf{Sim}  $\uparrow$
& \textbf{DNSMOS}  $\uparrow$ & \textbf{dWER}  $\downarrow$ & \textbf{Sim}  $\uparrow$
& \textbf{DNSMOS}  $\uparrow$ & \textbf{dWER}  $\downarrow$ & \textbf{Sim}  $\uparrow$
& \textbf{UTMOS}  $\uparrow$ & \textbf{dWER}  $\downarrow$ & \textbf{Sim}   $\uparrow$ \\
\midrule
Reference & --- & ---  
& 4.09 & 0.00 & 100.0
& 2.84 & 0.00 & 100.0
& 3.56 & 0.00 & 100.0
& 3.73 & 0.00 & 100.0
& 4.09 & 0.00 & 100.0 \\

EnCodec & 75.0 & 1.50 
& 1.58 & 8.08 & 93.8
& 1.33 & 29.60 & 95.5
& 2.76 & 28.16 & 87.7
& 2.40 & 55.17 & 86.3
& 1.24 & 86.52 & 72.2 \\

DAC & 50.0 & 1.00 
& 1.29 & 20.04 & 89.2
& 1.24 & 56.08 & 89.1
& 2.72 & 63.90 & 79.8
& 2.40 & 90.92 & 76.6
& 1.25 & 104.00 & 67.2 \\

WavLM6-KM & 50.0 & \underline{0.45} 
& 3.75 & 6.20 & 90.0
& \underline{2.97} & 44.54 & 89.5
& 3.06 & 20.67 & 82.9
& 2.87 & 36.60 & 85.9
& 2.90 & 26.68 & 92.4 \\

SpeechTokenizer & 50.0 & 1.00 
& 2.28 & 5.14 & 91.6
& 1.55 & 56.32 & 92.0
& 2.74 & 34.51 & 82.2
& 2.58 & 57.26 & 82.8
& 1.49 & \textbf{20.32} & 81.2 \\

SemantiCodec & 25.0 & 0.65 
& 2.91 & 8.97 & 96.0
& 1.87 & 36.21 & 97.7
& 3.13 & 31.46 & 90.6
& 2.67 & 51.18 & 89.9
& 2.02 & 106.00 & 72.8 \\

Mimi & 12.5 & 0.69 
& 3.29 & 5.73 & 96.0
& 2.08 & 30.96 & 96.7
& 3.01 & 28.00 & 87.8
& 2.65 & 49.14 & 89.4
& 2.40 & 110.00 & 89.7 \\

WavTokenizer & 40.0 & 0.48 
& 3.78 & 11.55 & 95.4
& 2.64 & 49.73 & 97.0
& 3.09 & 42.12 & 89.8
& 2.53 & 70.10 & 86.3
& 3.13 & 43.15 & 73.4 \\

BigCodec & 80.0 & 1.04 
& 4.11 & \underline{2.55} & \textbf{98.5}
& 2.86 & \underline{15.24} & \textbf{99.1}
& 3.19 & 20.67 & \textbf{92.3}
& 2.75 & 53.26 & 88.3
& 1.31 & 99.96 & 68.9 \\

Stable Codec & 25.0 & 0.70 
& \textbf{4.32} & 4.97 & 94.7
& \textbf{3.47} & 56.99 & 95.9
& \textbf{3.33} & 20.32 & 88.8
& 2.91 & 43.52 & 90.0
& \textbf{3.76} & 27.63 & 71.1 \\

{FocalCodec} & 50.0 & 0.65 
& \underline{4.05} & \textbf{2.18} & {97.4}
& 2.96 & \textbf{12.57} & 98.3
& 3.16 & \textbf{8.08} & {91.3}
& \textbf{2.93} & \textbf{27.89} & {91.6}
& \underline{3.38} & \underline{21.27} & \underline{92.2} \\



\midrule
\textbf{DyCAST-CA} & $\sim$14.4 & $\sim$0.92
& 3.99 & 3.32 & 97.4
& 2.85 & 16.89 & 98.3
& \underline{3.27} & 14.70 & 91.5
& \underline{2.92} & \underline{30.95} & \underline{91.8}
& 3.19 & 25.94 & 92.0 \\

\textbf{DyCAST-BP1} & $\sim$17.5 & $\sim$1.12
& 3.92 & 3.61 & \underline{97.6}
& 2.90 & 16.46 & \underline{98.5}
& 3.22 & \underline{12.60} & \underline{92.1}
& \underline{2.92} & 31.37 & \textbf{92.0}
& 3.26 & 24.77 & \textbf{92.4} \\

\textbf{DyCAST-BP3} & $\sim$\underline{9.0} & $\sim$0.57 
& 3.95 & 4.66 & 97.2
& 2.92 & 21.61 & 98.3
& 3.23 & 17.96 & 91.4
& 2.91 & 36.96 & 91.4
& 3.24 & 27.66 & 92.3 \\

\textbf{DyCAST-BP5} & $\sim$\textbf{6.2} & $\sim$\textbf{0.40}
& 3.97 & 8.84 & 96.5
& 2.94 & 34.04 & 97.7
& 3.22 & 26.62 & 89.7
& 2.90 & 46.42 & 90.0
& 3.21 & 35.59 & \underline{92.2} \\

\bottomrule
\end{tabular}
}
\end{footnotesize}
\end{center}
\vskip -0.15in
\end{table*}

\subsection{Speech Resynthesis and Voice Conversion}
We evaluate the speech resynthesis (SR) capabilities of DyCAST on \textbf{LibriSpeech}~\cite{panayotov2015librispeech}, \textbf{MLS}~\cite{pratap2020mls}, \textbf{VoiceBank}~\cite{valentinibotinhao2016voicebank}, and \textbf{Libri1Mix}. Naturalness, intelligibility, and speaker similarity are assessed using \textbf{UTMOS}~\cite{saeki2022utmos} for clean speech and \textbf{DNSMOS}~\cite{reddy2022dnsmos} for noisy speech, \textbf{dWER}~\cite{wang2021dwer}, and \textbf{Sim}~\cite{dellalibera2025focalcodec}, respectively. Additional details are provided in Appendix~\ref{subsec:appendix_sr}.
We further evaluate voice conversion (VC) on \textbf{VCTK}~\cite{veaux2017cstr}, following the protocol of \citet{dellalibera2025focalcodec}, which is described in detail in Appendix~\ref{subsec:appendix_vc}.

As shown in \cref{tab:speech_resynthesis}, DyCAST achieves a strong balance between reconstruction quality and token efficiency, operating at substantially lower frame rates than fixed-frame-rate baselines while maintaining competitive performance across all metrics.
On LibriSpeech, DyCAST variants consistently reach high naturalness scores and low intelligibility degradation, closely matching strong fixed-rate codecs such as FocalCodec and Stable Codec, despite using {3–8x fewer frames}. Notably, {DyCAST-CA} and {DyCAST-BP1} achieve dWER values comparable to high-rate baselines, confirming that character-aligned or predicted durations preserve linguistic content even at reduced token rates.
For {multilingual speech}, DyCAST maintains robust performance across languages, with only a moderate increase in dWER as the frame rate is reduced. Importantly, speaker similarity remains consistently high across all DyCAST variants, indicating that variable-rate tokenization does not compromise speaker identity, even in low-rate regimes.
Notably, aside from the character aligner, which naturally supports multilingual inputs, the codec itself is trained exclusively on English speech, highlighting the strong generalization capability of the learned discrete representations.
In noisy conditions, DyCAST remains competitive with dedicated speech codecs. While very aggressive compression leads to higher dWER, naturalness and speaker similarity remain stable, suggesting that explicit duration modeling provides robustness to noise by avoiding over-fragmentation of tokens.
For voice conversion, DyCAST achieves strong speaker similarity and intelligibility, comparable to FocalCodec. Performance degrades gracefully as the frame rate is reduced, indicating that the learned discrete units retain speaker-relevant information even at low temporal resolution.

\vspace{-0.2cm}
\paragraph{Retrieval-Augmented Decoding.}
To simulate a realistic deployment scenario, we construct a large candidate pool containing 32-dimensional \emph{continuous} latents extracted from the compressor. The pool includes all utterances from LibriSpeech \texttt{train-clean-100}, \texttt{dev-clean}, and \texttt{test-clean}, totaling approximately 20M vectors. This setting reflects a practical use case in which the retrieval database contains diverse speakers, utterances, and acoustic conditions, while also naturally including material from the current speaker, as speakers tend to repeat similar acoustic patterns over time. Although a speaker-specific pool would likely yield stronger results, this deliberately challenging setup allows us to assess the robustness of the learned latent space and its ability to retrieve relevant acoustic content from a large, heterogeneous database.
For efficiency, the retrieval pool is indexed using an inverted file (IVF) index~\cite{zobel2006ivf} implemented with the FAISS library~\cite{johnson2019billion}. We use $n_{\text{list}}=4096$ inverted lists, probe $n_{\text{probe}}=16$ lists at query time, and train the IVF index coarse clustering on a randomly subsampled set of 500k vectors.

\begin{table}[t!]
\setlength{\tabcolsep}{2pt}
\captionsetup{width=1.0\linewidth}
\caption{Retrieval-augmented decoding for different similarity thresholds $\tau$ on LibriSpeech resynthesis. \textbf{Best} and \underline{second-best} results are highlighted within each section.}
\vskip -0.1in
\label{tab:retrieval_ablation}
\centering
\begin{scriptsize}
\resizebox{0.75\linewidth}{!}{%
\begin{tabular}{ll|ccc}
\toprule
\multirow{1}{*}{\textbf{Codec}} &
\multirow{1}{*}{\textbf{Configuration}} &
\textbf{UTMOS} $\uparrow$ &
\textbf{dWER} $\downarrow$ &
\textbf{Sim} $\uparrow$ \\
\midrule

\multirow{4}{*}{\textbf{DyCAST-CA}}
& w/o retriever & \textbf{3.99} & 3.32 & \underline{97.4} \\
& retriever ($\tau=95$) & \underline{3.85} & \textbf{2.97} & \textbf{97.8} \\
& retriever ($\tau=97$) & 3.91 & \underline{3.00} & \textbf{97.8} \\
& retriever ($\tau=99$) & \textbf{3.99} & 3.33 & \underline{97.4} \\
\cmidrule{1-5}

\multirow{4}{*}{\textbf{DyCAST-BP1}}
& w/o retriever & \textbf{3.92} & 3.61 & \underline{97.6} \\
& retriever ($\tau=95$) & 3.80 & \underline{3.40} & \textbf{98.0} \\
& retriever ($\tau=97$) & \underline{3.84} & \textbf{3.36} & \textbf{98.0} \\
& retriever ($\tau=99$) & \textbf{3.92} & 3.77 & \underline{97.6} \\
\cmidrule{1-5}

\multirow{4}{*}{\textbf{DyCAST-BP3}}
& w/o retriever & \textbf{3.95} & 4.66 & 97.2 \\
& retriever ($\tau=95$) & 3.81 & \textbf{3.89} & \textbf{97.8} \\
& retriever ($\tau=97$) & 3.87 & 4.23 & \underline{97.7} \\
& retriever ($\tau=99$) & \underline{3.94} & 4.61 & 97.2 \\
\cmidrule{1-5}

\multirow{4}{*}{\textbf{DyCAST-BP5}}
& w/o retriever & \textbf{3.97} & 8.84 & 96.5 \\
& retriever ($\tau=95$) & 3.80 & \textbf{7.22} & \textbf{97.2} \\
& retriever ($\tau=97$) & 3.91 & 8.08 & 96.9 \\
& retriever ($\tau=99$) & \underline{3.96} & 8.87 & 96.5 \\

\bottomrule
\end{tabular}
}
\end{scriptsize}
\vskip -0.25in
\end{table}

\cref{tab:retrieval_ablation} evaluates the impact of retrieval-augmented decoding on LibriSpeech resynthesis across different similarity thresholds $\tau$. Overall, retrieval consistently improves intelligibility and speaker similarity at low token rates, while preserving naturalness when applied conservatively.
For all DyCAST variants, moderate similarity thresholds ($\tau \in [95,97]$) yield the most consistent gains. In this regime, retrieval reduces dWER relative to decoding without a retriever, with the largest improvements observed for lower-rate variants. This suggests that retrieval is particularly effective when token sequences are short and reconstruction ambiguity is higher, allowing the decoder to recover fine-grained acoustic detail without increasing bitrate.
Importantly, speaker similarity improves systematically with retrieval at moderate thresholds, indicating that retrieved exemplars reinforce speaker-specific characteristics rather than introducing identity drift.
At the same time, UTMOS remains stable or decreases only marginally, indicating that retrieval does not harm perceived naturalness when constrained by a sufficiently high similarity threshold. 
In contrast, very high thresholds ($\tau = 99$) closely match the no-retrieval baseline across all metrics, effectively disabling the mechanism by being overly selective.
We also note that UTMOS is a noisy metric and should be interpreted only as a coarse indicator of perceptual quality~\cite{dellalibera2025focalcodec}. Consistent with this observation, informal listening suggests that for UTMOS values above 3.50, similarity plays a more dominant role in perceived quality, with even small improvements corresponding to audible gains.

\vspace{-0.1cm}
\subsection{Discriminative Tasks}
We evaluate the representational quality of DyCAST discrete tokens on automatic speech recognition (ASR), speaker identification (SI), and speech emotion recognition (SER) probing tasks~\cite{mousavi2024dasb}. For ASR and SI, we use \textbf{LibriSpeech}, while for SER we use \textbf{IEMOCAP}~\cite{busso2008iemocap}. All experiments follow the same shallow probing setup as in \cite{dellalibera2025focalcodec}. We report word error rate (\textbf{WER}) for ASR and error rate (\textbf{ER}) for SI and SER. Additional details are provided in Appendix~\ref{subsec:appendix_asr}--\ref{subsec:appendix_ser}.

Results for discriminative probing tasks are reported in \cref{tab:downstream_tts}. Overall, DyCAST performs well despite operating at substantially lower frame rates than fixed-rate baselines. On ASR, DyCAST-CA achieves the best WER among all codecs, outperforming both fixed-frame-rate speech codecs and prior discrete tokenizers, while using significantly fewer tokens. This highlights the benefit of character-aligned tokenization for preserving fine-grained linguistic information that is directly relevant for recognition.
As the frame rate is reduced further, performance degrades gracefully. Variants relying on predicted boundaries maintain competitive WER at moderate rates, while extremely low-frame-rate configurations exhibit higher WER, reflecting the expected trade-off between compression and phonetic resolution. Importantly, this degradation is consistent with trends observed in fixed-rate codecs at comparable or higher bitrates.
For SI, DyCAST achieves error rates comparable to strong baselines, indicating that speaker-related cues are well preserved even under aggressive temporal compression. Although fixed-frame-rate tokenizers optimized for speaker modeling (\eg WavTokenizer) achieve lower SI error rates, DyCAST offers a more favorable balance between speaker information retention and token efficiency.
On SER, performance across all codecs remains relatively close, suggesting that coarse prosodic and affective cues are robust to different tokenization strategies. DyCAST performs on par with fixed-rate baselines, indicating that variable-rate tokenization does not impair the extraction of emotional content, despite operating at lower frame rates.
Overall, these results indicate that DyCAST tokens retain rich linguistic, speaker, and paralinguistic information even at substantially reduced rates. In particular, the strong ASR performance of the character-aligned variant suggests that explicitly modeling durations and linguistic alignment yields discrete tokens that are well suited for speech understanding tasks.

\subsection{Text-To-Speech}
We evaluate the generative capabilities of DyCAST on text-to-speech (TTS) using \textbf{LibriSpeech}, with character-level text as input and speaker embeddings for speaker conditioning. Performance is assessed using the same metrics as for speech resynthesis, namely \textbf{UTMOS}, \textbf{dWER}, and \textbf{Sim}.
Following \citet{dellalibera2025focalcodec}, we employ an autoregressive architecture over speech tokens for TTS, reflecting the formulation of TTS as text-conditioned speech language modeling. The only exception is the {DyCAST-CA} variant: thanks to the \emph{hard} character-level alignment provided by the tokenizer, characters and speech tokens are in one-to-one correspondence. This enables the use of a \textbf{non-autoregressive TTS} model that directly maps each input character to its corresponding speech token.
Details on the model architecture, hyperparameters, and training procedure are provided in Appendix~\ref{subsec:appendix_tts}.

As shown in \cref{tab:downstream_tts}, DyCAST achieves strong performance across naturalness, intelligibility, and speaker similarity while operating at substantially reduced frame rates. Since the TTS model is autoregressive, lower frame rates directly translate into shorter sequences, yielding a more favorable learning regime by reducing exposure bias and easing long-range dependency modeling.
Notably, DyCAST-CA clearly stands out, achieving by far the best TTS performance across all metrics. This result is enabled by the non-autoregressive one-to-one architecture uniquely supported by DyCAST-CA, which eliminates sequential token prediction altogether and is particularly effective in this limited-data regime. In addition, this architecture enables extremely fast inference, as generation is performed in a single parallel pass and does not require sampling or rescoring multiple hypotheses.
Overall, these results highlight the dual benefit of DyCAST: efficient, controllable, low-rate tokenization that simplifies autoregressive generation, and character-aligned representations that enable state-of-the-art quality and high inference efficiency in data-limited TTS scenarios.

\begin{table}[t!]
\setlength{\tabcolsep}{2pt}
\caption{Downstream modeling performance. Shallow probing heads are used for discriminative tasks (ASR -- LibriSpeech, SI -- LibriSpeech, SER -- IEMOCAP), transformer-based models are used for generative tasks (TTS -- LibriSpeech). \textbf{Best} and \underline{second-best} results are highlighted.}
\vskip -0.1in
\label{tab:downstream_tts}
\centering

\begin{threeparttable}
\begin{scriptsize}
\resizebox{1.0\linewidth}{!}{%
\begin{tabular}{l c c|c|c|c|c c c}
\toprule
\multirow{3}{*}{\textbf{Codec}} &
\multirow{3}{*}{\makecell{\textbf{Frame} \\ \textbf{Rate (Hz)}} $\downarrow$} &
\multirow{3}{*}{\makecell{\textbf{Bitrate} \\ \textbf{(kbps)}} $\downarrow$} &
\textbf{ASR} &
\textbf{SI} &
\textbf{SER} &
\multicolumn{3}{c}{\textbf{TTS}} \\
\cmidrule{4-9}
& & &
\textbf{WER} $\downarrow$ &
\textbf{ER} $\downarrow$ &
\textbf{ER} $\downarrow$ &
\textbf{UTMOS} $\uparrow$ &
\textbf{dWER} $\downarrow$ &
\textbf{Sim} $\uparrow$ \\
\midrule

EnCodec & 75.0 & 1.50 & 27.89 & 3.00 & 47.00 & 1.71 & 64.28 & 83.2 \\
DAC & 50.0 & 1.00 & 35.89 & 3.27 & 45.90 & 1.34 & 47.06 & 85.9 \\
WavLM6-KM & 50.0 & \underline{0.45} & 19.04 & 22.30 & \underline{42.90} & 3.74 & 38.67 & 88.7 \\
SpeechTokenizer & 50.0 & 1.00 & \underline{14.97} & 2.73 & \textbf{41.50} & 2.69 & 35.46 & 89.2 \\
SemantiCodec & 25.0 & 0.65 & 41.42 & 15.90 & 51.60 & 2.82 & 48.38 & 91.4 \\
Mimi & 12.5 & 0.69 & 22.98 & 5.43 & 44.70 & 3.11 & 28.63 & \underline{93.6} \\
WavTokenizer & 40.0 & 0.48 & 35.62 & \underline{2.44} & 49.80 & 3.68 & 47.56 & 92.8 \\
BigCodec & 80.0 & 1.04 & 26.41 & \textbf{2.34} & 47.50 & 3.43 & 54.43 & 89.4 \\
Stable Codec & 25.0 & 0.70 & 16.85 & 16.50 & 46.54 & 3.19 & 49.28 & 88.8 \\
FocalCodec & 50.0 & 0.65 & 17.63 & 4.48 & 45.60 & \underline{4.11} & 28.10 & {93.3} \\

\midrule
\textbf{DyCAST-CA} & $\sim$14.4 & $\sim$0.92 & \textbf{13.05} & 9.03 & 49.54 &
\textbf{4.20}\tnote{$\dagger$} & \textbf{3.42}\tnote{$\dagger$} & \textbf{96.2}\tnote{$\dagger$} \\
\textbf{DyCAST-BP1} & $\sim$17.5 & $\sim$1.12 & 19.98 & 7.34 & 51.61 & 3.91 & \underline{11.57} & 93.2 \\
\textbf{DyCAST-BP3} & $\sim$\underline{9.0} & $\sim$0.57  & 27.80 & 7.42 & 50.00 & 4.00 & 14.56 & 92.6 \\
\textbf{DyCAST-BP5} & $\sim$\textbf{6.2} & $\sim$\textbf{0.40} & 40.27 & 11.50 & 50.23 & 4.02 & 15.85 & 92.8 \\

\bottomrule
\end{tabular}
}
\end{scriptsize}

\begin{tablenotes}[flushleft]
\scriptsize
\vspace{-0.5mm}
\item[\textnormal{$\dagger$}] \hspace{-1mm} Non-autoregressive one-to-one architecture, defined only for {DyCAST-CA}.
\end{tablenotes}
\end{threeparttable}

\vspace{-0.3in}
\end{table}

\section{Conclusions}
We introduced DyCAST, a speech codec that enables variable-frame-rate tokenization through soft character-level alignment and explicit duration modeling. DyCAST produces substantially shorter token sequences than fixed-frame-rate codecs while maintaining competitive speech resynthesis quality and downstream performance, resulting in improved efficiency and more tractable transformer-based sequence modeling.
A key advantage of DyCAST is its flexibility, enabling content-adaptive frame rates and explicit control over duration, bitrate, and reconstruction quality. Combined with the proposed retrieval-augmented decoding strategy for improved resynthesis, DyCAST offers a promising foundation for next-generation speech tokenization.





\bibliography{bibliography}
\bibliographystyle{icml2026}

\newpage
\appendix
\onecolumn

\section{Datasets}
\label{sec:appendix_datasets}
The following datasets were used in this work:
\begin{itemize}[topsep=0pt, leftmargin=15pt]
    \item \textbf{LibriSpeech}~\cite{panayotov2015librispeech} is a large-scale corpus of English read speech derived from audiobooks in the LibriVox project. It contains approximately 1000 hours of speech sampled at 16 kHz, with predefined training, validation, and test splits. License: CC BY 4.0.

    \item \textbf{LibriTTS}~\cite{zen2019libritts} is a corpus designed for text-to-speech research, constructed from the same source as LibriSpeech. It consists of 585 hours of transcribed speech with predefined training, validation, and test splits. License: CC BY 4.0.

    \item \textbf{MLS}~\cite{pratap2020mls} is an extension of LibriSpeech to multiple languages, including English, German, Dutch, French, Spanish, Italian, Portuguese and Polish. It provides approximately 44,500 hours of transcribed English speech and about 6000 hours from other languages. License: CC BY 4.0.

    \item \textbf{VoiceBank}~\cite{valentinibotinhao2016voicebank} is a dataset primarily used for speech enhancement, including 11,572 utterances from 28 speakers in the training set (noise at 0 dB, 5 dB, 10 dB, and 15 dB), and 872 utterances from 2 unseen speakers in the test set (noise at 2.5 dB, 7.5 dB, 12.5 dB, and 17.5 dB). License: CC BY 4.0.

    \item \textbf{LibriMix}~\cite{cosentino2020librimix} is a dataset for speech separation and enhancement, created by mixing LibriSpeech utterances with noise from the WHAM!~\cite{wichern2019wham} corpus. It provides mixtures of two or three speakers at different signal-to-noise ratios. License: MIT.

    \item \textbf{VCTK}~\cite{veaux2017cstr} is a corpus of English speech recordings from 110 speakers with various accents. It is widely used for speaker adaptation, text-to-speech, and voice conversion tasks. License: CC BY 4.0.

    \item \textbf{IEMOCAP}~\cite{busso2008iemocap} is a dataset designed for emotion recognition, consisting of scripted and improvised dialogues performed by 10 actors. It includes audio, video, and textual transcriptions with emotion labels such as happiness, sadness, and anger. License: \href{https://sail.usc.edu/iemocap/iemocap\_release.htm}{https://sail.usc.edu/iemocap/iemocap\_release.htm}.
\end{itemize}

\section{Baselines}
\label{sec:appendix_baselines_codecs}
We compare DyCAST against a diverse set of fixed-frame-rate neural audio codecs covering general audio, speech-focused, single- and multi-codebook, acoustic, semantic, and hybrid (additional details are provided in \cref{tab:appendix_baselines_codecs}):
\begin{itemize}[topsep=0pt, leftmargin=15pt]
    \item \textbf{EnCodec}~\cite{defossez2023encodec}: A general-purpose neural audio codec supporting both causal and non-causal operation, trained on large-scale heterogeneous audio data and widely used as a strong baseline for speech and audio compression.
    \item \textbf{DAC}~\cite{kumar2023dac}: A high-quality non-causal audio codec optimized for perceptual reconstruction, trained on diverse speech and music datasets and commonly used in downstream speech tasks.
    \item \textbf{WavLM6-KM}~\cite{wang2024selm}: A speech codec obtained by clsutering WavLM representations from the 6-th transfomer layer.
    \item \textbf{SpeechTokenizer}~\cite{zhang2024speechtokenizer}: A multi-codebook speech tokenizer designed to disentangle content and speaker information across codebooks, primarily trained on LibriSpeech.
    \item \textbf{SemantiCodec}~\cite{liu2024semanticodec}: A general audio codec trained on large-scale multimodal datasets, explicitly targeting semantic preservation across speech and non-speech audio.
    \item \textbf{Mimi}~\cite{defossez2024moshi}: A large-scale causal speech tokenizer trained on millions of hours of speech, designed for streaming and speech language modeling applications.
    \item \textbf{WavTokenizer}~\cite{ji2024wavtokenizer}: A unified audio tokenizer trained on mixed speech and music data, supporting multilingual and multi-domain speech modeling.
    \item \textbf{BigCodec}~\cite{xin2024bigcodec}: A speech-focused codec trained on LibriSpeech, emphasizing efficient discrete representations for downstream speech generation and modeling tasks.
    \item \textbf{Stable Codec}~\cite{parker2024scaling}: A transformer-based neural speech codec trained on large-scale speech corpora, supporting optional causal inference.
    \item \textbf{FocalCodec}~\cite{dellalibera2025focalcodec}: A low-bitrate speech codec based on focal modulation and binary spherical quantization, serving as the closest architectural baseline to DyCAST.
\end{itemize}

\begin{table*}[t!]
\setlength{\tabcolsep}{3pt}
\caption{Baseline codecs.}
\vskip -0.15in
\label{tab:appendix_baselines_codecs}
\begin{center}
\resizebox{\textwidth}{!}{%
\begin{tabular}{lcccccc}
\toprule
\textbf{Codec} & \textbf{Training Datasets} & \textbf{Hours} & \textbf{Multilingual} & \textbf{Audio Domain} & \textbf{Checkpoint} & \textbf{License} \\
\midrule
\multirow{2}{*}{EnCodec~\cite{defossez2023encodec}} & \multirow{2}{*}{DNS, CommonVoice, AudioSet, FSD50K, Jamendo} & \multirow{2}{*}{17k+} & \multirow{2}{*}{Yes} & \multirow{2}{*}{General} & \multirow{2}{*}{\href{https://huggingface.co/facebook/encodec_24khz}{encodec\_24khz}} & \multirow{2}{*}{MIT} \\
& & & & & & \\
\multirow{2}{*}{DAC~\cite{kumar2023dac}} & \multirow{2}{*}{DAPS, DNS, CommonVoice, VCTK, MUSDB, Jamendo} & \multirow{2}{*}{10k+} & \multirow{2}{*}{Yes} & \multirow{2}{*}{General} & \multirow{2}{*}{\href{https://github.com/descriptinc/descript-audio-codec/releases/download/0.0.5/weights_16khz.pth}{weights\_16khz.pth}} & \multirow{2}{*}{MIT} \\
& & & & & & \\
\multirow{3}{*}{WavLM6-KM~\cite{wang2024selm}} & \multirow{3}{*}{\makecell{Subset of LibriSpeech (in addition to Libri-Light, \\ GigaSpeech, and VoxPopuli English for WavLM pretraining)}} & \multirow{3}{*}{\makecell{460 \\ (+ 94k)}} & \multirow{3}{*}{No} & \multirow{3}{*}{Speech} & \multirow{3}{*}{\href{https://huggingface.co/lucadellalib/discrete-wavlm-codec}{discrete-wavlm-codec}} & \multirow{3}{*}{Apache 2.0} \\
& & & & & & \\
& & & & & & \\
\multirow{2}{*}{SpeechTokenizer~\cite{zhang2024speechtokenizer}} & \multirow{2}{*}{LibriSpeech} & \multirow{2}{*}{960} & \multirow{2}{*}{No} & \multirow{2}{*}{Speech} & \multirow{2}{*}{\href{https://huggingface.co/fnlp/SpeechTokenizer/tree/main/speechtokenizer_hubert_avg}{speechtokenizer\_hubert\_avg}} & \multirow{2}{*}{Apache 2.0} \\
& & & & & & \\
\multirow{2}{*}{SemantiCodec~\cite{liu2024semanticodec}} & \multirow{2}{*}{\makecell{GigaSpeech, subset of OpenSLR, Million Song Dataset, \\ MedleyDB, MUSDB18, AudioSet, WavCaps, VGGSound}} & \multirow{2}{*}{20k+} & \multirow{2}{*}{Yes} & \multirow{2}{*}{General} & \multirow{2}{*}{\href{https://huggingface.co/haoheliu/SemantiCodec/tree/main/semanticodec_tokenrate_50}{semanticodec\_tokenrate\_50}} & \multirow{2}{*}{MIT} \\
& & & & & & \\
\multirow{3}{*}{Mimi~\cite{defossez2024moshi}} & \multirow{3}{*}{\makecell{Predominantly English speech (in addition to Libri-Light, \\ GigaSpeech, and VoxPopuli English for WavLM pretraining)}} & \multirow{3}{*}{\makecell{7M \\ (+ 94k)}} & \multirow{3}{*}{Likely} & \multirow{3}{*}{Speech} & \multirow{3}{*}{\href{https://huggingface.co/kyutai/mimi}{mimi}} & \multirow{3}{*}{CC BY 4.0} \\
& & & & & & \\
& & & & & & \\
\multirow{2}{*}{WavTokenizer~\cite{ji2024wavtokenizer}} & \multirow{2}{*}{\makecell{LibriTTS, VCTK, subset of CommonVoice, \\ subset of AudioSet, Jamendo, MUSDB}} & \multirow{2}{*}{8k} & \multirow{2}{*}{Yes} & \multirow{2}{*}{General} & \multirow{2}{*}{\href{https://huggingface.co/novateur/WavTokenizer-large-unify-40token}{WavTokenizer-large-unify-40token}} & \multirow{2}{*}{MIT} \\
& & & & & & \\
\multirow{2}{*}{BigCodec~\cite{xin2024bigcodec}} & \multirow{2}{*}{LibriSpeech} & \multirow{2}{*}{960} & \multirow{2}{*}{No} & \multirow{2}{*}{Speech} & \multirow{2}{*}{\href{https://huggingface.co/Alethia/BigCodec/resolve/main/bigcodec.pt}{bigcodec.pt}} & \multirow{2}{*}{MIT} \\
& & & & & & \\
\multirow{2}{*}{Stable Codec~\cite{parker2024scaling}} & \multirow{2}{*}{Libri-Light, MLS English} & \multirow{2}{*}{105k} & \multirow{2}{*}{No} & \multirow{2}{*}{Speech} & \multirow{2}{*}{\href{https://huggingface.co/stabilityai/stable-codec-speech-16k}{stable-codec-speech-16k}} & \multirow{2}{*}{\href{https://huggingface.co/stabilityai/stable-codec-speech-16k/blob/main/LICENSE.md}{StabilityAI}} \\
& & & & & & \\
\multirow{2}{*}{FocalCodec~\cite{dellalibera2025focalcodec}} & \multirow{2}{*}{\makecell{LibriTTS (in addition to Libri-Light, \\ GigaSpeech, and VoxPopuli English for WavLM pretraining)}} & \multirow{2}{*}{\makecell{585 \\ (+ 94k)}} & \multirow{2}{*}{No} & \multirow{2}{*}{Speech} & \multirow{2}{*}{\href{https://huggingface.co/lucadellalib/focalcodec_50hz}{focalcodec\_50hz}} & \multirow{2}{*}{Apache 2.0} \\
& & & & & & \\
\bottomrule
\end{tabular}
}
\end{center}
\vspace{-0.5cm}
\end{table*}

\section{Hyperparameters and Training Details}
\label{sec:hyperparameters}
We summarize the main hyperparameters and training settings used for all DyCAST components. All learnable modules are trained on LibriTTS~\cite{zen2019libritts} (585 hours from 2456 speakers) using full utterances and identical optimizer settings, except for the decoder, which requires a different training setup.
For all other modules, training is performed using the AdamW~\cite{loshchilov2019adamw} optimizer with an initial learning rate of 0.0005, $\beta_1 = 0.8$, $\beta_2 = 0.99$, and a weight decay of 0.01. The learning rate is reduced by a factor of 0.9 when the validation loss does not improve by at least 0.0025. Gradients are clipped to a maximum $L_2$ norm of 5, and training stops when the validation loss fails to decrease for several consecutive epochs.

\paragraph{Character Aligner.}
As a character aligner, we use MMS~\cite{pratap2024mms}, a 1B-parameter multilingual wav2vec2.0 ASR model pretrained on approximately 500k hours of speech across more than 1,000 languages and equipped with a CTC-based character prediction head. The model is kept frozen throughout training and is used to obtain character-level durations via simple \texttt{argmax} decoding. While more sophisticated alignment strategies such as Viterbi decoding~\cite{graves2006connectionist} are possible, we adopt \texttt{argmax} decoding for its simplicity and efficiency.

\paragraph{Encoder.}
For the encoder, we use the pretrained WavLM-large\footnote{\href{https://github.com/microsoft/unilm/tree/master/wavlm}{https://github.com/microsoft/unilm/tree/master/wavlm}} model~\cite{chen2022wavlm}. In particular, we extract representations from the 6th transformer layer, which provide semantically rich features while retaining fine-grained acoustic information~\cite{dellalibera2025focalcodec, baas2023knnvc}. The encoder is kept frozen throughout training.

\paragraph{Compressor.}
The compressor takes as input 1024-dimensional WavLM features and processes them through 3 focal downscaling blocks~\cite{dellalibera2025focalcodec} while preserving the original temporal resolution (50 Hz), each with a hidden dimension of 1024. Each block uses 2 focal levels, a window size of 14, a focal factor of 4, and a layer scale initialization of 0.0001. A final projection maps the 1024-dimensional hidden states to 32-dimensional latent representations.

\paragraph{Chunker.}
Within the chunker, the boundary predictor adopts the same architecture as the compressor. A final binary classification head maps the 1024-dimensional hidden states to boundary logits. The dynamic downsampling operation itself is implemented as a simple frame-selection and does not introduce additional trainable parameters.

\paragraph{Quantizer.}
The scalar spherical quantizer discretizes the continuous 32-dimensional latent representations using 4 levels per dimension, resulting in an implicit codebook of size $|\mathcal{C}| = 4^{32}$. In practice, it is infeasible to enumerate all codes explicitly; instead, we represent tokens by returning per-dimension indices in $\{0,1,2,3\}$. The quantizer has no trainable parameters, and we set the entropy loss weight to 0.1.

\paragraph{Dechunker.}
Within the dechunker, the duration predictor follows the same architectural design but processes 32-dimensional pooled quantized latents. The model predicts a free mean parameter (a scalar value) together with a shared, learnable scalar dispersion parameter $\alpha$. Dynamic upsampling is implemented by frame repetition and introduces no additional parameters; the length regularization strength is set to $\lambda = 0.05$.

\paragraph{Decompressor.}
The decompressor mirrors the compressor architecture, replacing focal downscaling blocks with focal upscaling blocks to reconstruct 1024-dimensional continuous representations from the quantized latent codes. 

\paragraph{Decoder.}
The Vocos~\cite{siuzdak2023vocos} decoder operates on 1024-dimensional WavLM features and processes them through 8 ConvNeXt blocks with a hidden dimension of 512, a feed-forward dimension of 1536, a kernel size of 7, and padding of 3. For the STFT, we use an FFT size of 1024 samples and a hop length of 320. The feature-matching loss is computed using 80-dimensional log-Mel spectrograms with the same STFT configuration. The discriminator follows the convolutional architecture introduced in~\cite{kong2020hifigan}.
Training is conducted on LibriTTS \texttt{train-clean-100} using audio chunks of 7040 samples and a batch size of 16. We use the AdamW optimizer with an initial learning rate of 0.0002, $\beta_1$ of 0.8, $\beta_2$ of 0.99, and a weight decay of 0.01. The learning rate follows an exponential decay schedule with a factor of 0.999. Training continues until perceived audio quality saturates, which occurs after approximately 3M steps.

\section{Downstream Evaluation}
\label{subsec:appendix_downstream}
In this section, we outline the downstream tasks used to evaluate the proposed codec and describe the corresponding experimental setups. Unless otherwise stated, all downstream models are trained using the AdamW optimizer~\cite{loshchilov2019adamw} with a batch size of 16, an initial learning rate of 0.0001, $\beta_1$ = 0.8, $\beta_2$ = 0.99, weight decay of 0.01, and dropout of 0.1. The learning rate is reduced by a factor of 0.9 if validation loss does not improve within a margin of 0.0025. Gradients are clipped to a maximum $L_2$ norm of 0.01. Training stops if validation loss does not decrease for several consecutive epochs. Each model is trained on a single NVIDIA H100 GPU with 80 GB of memory. Software for the downstream evaluation was implemented in Python using the SpeechBrain~\cite{ravanelli2021speechbrain, ravanelli2024open} toolkit.

\subsection{Speech Resynthesis (SR)}
\label{subsec:appendix_sr}
This task evaluates the ability of the codec to reconstruct high-quality speech from discrete tokens while preserving naturalness, speaker identity, and intelligibility.
For English, we use the \textbf{LibriSpeech}~\cite{panayotov2015librispeech} \texttt{test-clean} split. For multilingual evaluation, we use the dataset constructed by \citet{dellalibera2025focalcodec}, which was built by selecting 100 utterances from each of the seven non-English languages in \textbf{MLS}~\cite{pratap2020mls} (Dutch, French, German, Italian, Polish, Portuguese, and Spanish) for a total of 700 utterances\footnote{\href{https://zenodo.org/records/14791114}{https://zenodo.org/records/14791114}}.
To assess robustness to environmental noise, we additionally evaluate on the test sets of \textbf{VoiceBank}~\cite{valentinibotinhao2016voicebank} and \textbf{Libri1Mix}, constructed by mixing clean utterances from the first speaker of LibriMix~\cite{cosentino2020librimix} with noise from WHAM!~\cite{wichern2019wham}. Naturalness is measured using \textbf{UTMOS}~\cite{saeki2022utmos} for clean speech and \textbf{DNSMOS}~\cite{reddy2022dnsmos} for noisy speech. Intelligibility is evaluated using differential word error rate (\textbf{dWER})~\cite{wang2021dwer}, computed from transcriptions obtained with Whisper-small~\cite{radford2022robust}\footnote{\href{https://huggingface.co/openai/whisper-small}{https://huggingface.co/openai/whisper-small}}. Speaker fidelity is assessed via cosine similarity (\textbf{Sim}) between speaker embeddings extracted using WavLM-base-SV~\cite{chen2022wavlm}\footnote{\href{https://huggingface.co/microsoft/wavlm-base-sv}{https://huggingface.co/microsoft/wavlm-base-sv}}. We do not report signal-level metrics such as SNR, PESQ~\cite{rix2001pesq}, or STOI~\cite{taal2011stoi}, as they correlate poorly with perceived reconstruction quality~\cite{parker2024scaling, wang2024selm, dellalibera2025focalcodec}, and we avoid stronger ASR models to prevent masking reconstruction artifacts.

\subsection{Voice Conversion (VC)}
\label{subsec:appendix_vc}
This task evaluates the ability of the codec to disentangle speaker identity from linguistic content despite its single-codebook design. Voice conversion is performed by converting speech from a source speaker to a target speaker using reference speech from the target speaker. 
For single-codebook baselines, including DyCAST, we follow the $k$-nearest neighbors (kNN) approach of \citet{baas2023knnvc}. Specifically, each frame in the reconstructed feature sequence (immediately before the decoder) is replaced by the average of the $k=4$ nearest neighbors (measured by cosine similarity) from continuous features extracted from the target speaker reference utterance. 
For multi-codebook baselines, we follow the procedure of \citet{zhang2024speechtokenizer}. Both source and reference speech are tokenized, and voice conversion is performed by concatenating the first codebook tokens from the source with the second-to-last codebook tokens from the reference, before decoding. If sequence lengths differ, the reference sequence is truncated or circularly padded as needed. Effective disentanglement of content and speaker information across codebooks is expected to yield strong voice conversion performance.
We conduct voice conversion experiments on \textbf{VCTK}~\cite{veaux2017cstr}, which contains parallel utterances from multiple speakers. For each sample, we randomly select (i) an utterance from a source speaker, (ii) the corresponding parallel utterance from a target speaker, and (iii) a different utterance from the same target speaker to serve as the reference. Among candidate reference utterances, we select the longest to minimize padding. Repeating this procedure for each speaker and each of the approximately 24 parallel utterances yields a total of 2,521 evaluation samples. 
Performance is evaluated using the same metrics as in speech resynthesis: \textbf{UTMOS}, \textbf{dWER}, and \textbf{Sim}.

\subsection{Automatic Speech Recognition (ASR)}
\label{subsec:appendix_asr}
This task evaluates the quality of the learned discrete representations through a probing setup, where a shallow model is trained on top of fixed codec tokens to assess their linguistic content. We use \textbf{LibriSpeech}~\cite{panayotov2015librispeech} \texttt{train-clean-100} and \texttt{train-clean-360} for training, \texttt{dev-clean} for validation, and \texttt{test-clean} for evaluation.
The ASR probe consists of a 2-layer bidirectional LSTM with 512-dimensional hidden states, followed by a CTC~\cite{graves2006connectionist} prediction head. The model is trained to predict either characters or byte-pair encoding (BPE) units, using BPE vocabularies of size 1000, except for Mimi, which achieves its best performance with a vocabulary size of 500. When a codec employs multiple codebooks, we compute a learned weighted sum of the embeddings from each codebook, following~\cite{chen2022wavlm}. The embedding layer is initialized using the discrete embeddings from the codec quantizer. Performance is reported in terms of word error rate (\textbf{WER}).

\subsection{Speaker Identification (SI)}
\label{subsec:appendix_si}
This task probes whether the learned discrete representations encode speaker identity through a probing setup, where a shallow model is trained on top of fixed codec tokens to assess their acoustic content.
We use \textbf{LibriSpeech}~\cite{panayotov2015librispeech}, grouping utterances from \texttt{train-clean-100} and \texttt{train-clean-360} by speaker ID. The data are randomly split into training, validation, and test sets with a ratio of 80\% / 10\% / 10\%.
The SI probe closely mirrors the ASR setup. A 2-layer bidirectional LSTM with 512-dimensional hidden states processes the token sequence, and the output is aggregated using statistics pooling, followed by a cross-entropy classification head. When multiple codebooks are present, embeddings are combined via a learned weighted sum, as in ASR. Performance is reported in terms of speaker error rate (\textbf{ER}).

\subsection{Speech Emotion Recognition (SER)}
\label{subsec:appendix_ser}
This task probes whether the learned representations capture paralinguistic information related to emotion. We again adopt a probing setup with a shallow classifier trained on top of frozen codec tokens. We use the \textbf{IEMOCAP} dataset~\cite{busso2008iemocap}, focusing on four emotion classes: sadness, happiness, anger, and neutral. Sessions 1--4 are used for training, session 5F for validation, and session 5M for testing.
The SER setup is identical to SI, with the only difference being the number of output classes in the classification head. Performance is reported in terms of emotion error rate (\textbf{ER}).

\subsection{Text-To-Speech (TTS)}
\label{subsec:appendix_tts}
This task evaluates the suitability of discrete speech representations for speech generation from text. We use \textbf{LibriSpeech} \texttt{train-clean-100} and \texttt{train-clean-360} for training, \texttt{dev-clean} for validation, and \texttt{test-clean} for evaluation. The \texttt{test-clean} split contains a small fraction of long utterances (approximately 4\%) exceeding 20 seconds, which are largely absent from the training data; to reduce this train-test mismatch, we remove these long utterances from the test set. The model input consists of character-level text tokens, while the target sequence comprises speech tokens extracted from the corresponding utterances.
We consider two TTS model variants. An \textbf{autoregressive} model is used for all codecs except DyCAST-CA, while a \textbf{non-autoregressive} model is used exclusively for DyCAST-CA to exploit the availability of character-aligned tokens. Performance is evaluated using \textbf{UTMOS}, \textbf{dWER}, and \textbf{Sim}.

\vspace{-0.09cm}

\paragraph{Autoregressive TTS.} The autoregressive model is a Llama~3 decoder~\cite{dubey2024llama3herdmodels} with 12 layers, 4 attention heads, 1 key-value head, a model dimension of 512, a feed-forward dimension of 2048, and a base RoPE frequency of 10{,}000. Speaker identity is provided by extracting embeddings from the target utterance using WavLM-base~\cite{chen2022wavlm}, fine-tuned for speaker verification. The pooled speaker embedding is prepended to the character embeddings to condition the model on speaker identity. The embedding layer is initialized using the discrete embeddings from the codec quantizer.
Training is performed using next-token prediction, where the input sequence consists of the pooled speaker embedding, character embeddings, and speech token embeddings. The cross-entropy loss is computed only over speech tokens, while character and speaker embeddings are excluded from the loss.
If the codec has multiple codebooks, we flatten tokens across the codebook dimension, as this approach has been shown to provide the best downstream performance given a sufficiently high computational budget~\cite{copet2023musicgen}.
At inference time, we use top-$p$ sampling with $\beta_1$ = 0.9 and a temperature of 1.0. Following~\citet{tian2025espnetslm,dellalibera2025focalcodec}, we generate five samples per utterance and select the one with the lowest word error rate relative to the input text, computed using Whisper-small~\cite{radford2022robust}.

\paragraph{Non-autoregressive TTS.} The non-autoregressive  model employs the same Llama~3 architecture, where causal self-attention is replaced with bidirectional self-attention, while all other components remain unchanged. The pooled speaker embedding is replicated for the number of character positions and summed with the hidden representations at each layer to condition the model on speaker identity. Training is performed using per-character target prediction with a cross-entropy loss at each step. At inference time, decoding is performed using \texttt{argmax}.

\section{Additional Results}
\label{sec:additional_results}

\subsection{Decoding Mode}
\label{sec:appendix_decode_mode}
\cref{tab:duration_ablation} analyzes the impact of different decoding modes on LibriSpeech speech resynthesis. Overall, explicitly providing duration information consistently improves intelligibility, while naturalness and speaker similarity remain largely stable across decoding modes.
For all DyCAST variants, \emph{tokens + durations} yields the lowest dWER, confirming that explicit duration supervision enables more accurate temporal reconstruction and reduces linguistic distortions. This effect is particularly pronounced for predicted-duration variants, where dWER increases moderately as duration information is removed, highlighting the importance of duration modeling at low token rates.
In contrast, \emph{tokens only} decoding achieves comparable or slightly higher UTMOS across all variants, indicating that naturalness is largely preserved even without explicit timing cues. However, this comes at the cost of degraded intelligibility, especially for aggressive compression regimes, where the absence of duration constraints leads to temporal misalignment and increased recognition errors.
The intermediate \emph{tokens + utterance length} mode offers a favorable trade-off, recovering most of the intelligibility gains of full duration decoding while requiring only a global length constraint. Notably, this mode performs robustly across all variants and is used as the default setting in our experiments unless otherwise stated.
Overall, these results show that DyCAST enables flexible decoding strategies with explicit control over the trade-off between timing accuracy, intelligibility, and decoding complexity, further emphasizing the practical advantages of variable-rate tokenization with explicit duration modeling.

\vskip -0.4cm
\begin{table*}[h!]
\setlength{\tabcolsep}{2pt}
\captionsetup{width=0.85\linewidth}
\caption{Effect of different decoding modes on LibriSpeech resynthesis.}
\vskip -0.1in
\label{tab:duration_ablation}
\centering
\resizebox{0.60\linewidth}{!}{%
\begin{tabular}{ll|ccc}
\toprule
\multirow{1}{*}{\textbf{Codec}} &
\multirow{1}{*}{\textbf{Decoding Mode}} &
\textbf{UTMOS} $\uparrow$ &
\textbf{dWER} $\downarrow$ &
\textbf{Sim} $\uparrow$ \\
\midrule

\multirow{3}{*}{\textbf{DyCAST-CA}}
& {Tokens + durations} & 3.85 & \textbf{3.07} & \textbf{97.5} \\
& {Tokens + utterance length} & \underline{3.99} & \underline{3.32} & \underline{97.4} \\
& {Tokens only} & \textbf{4.03} & 3.41 & \underline{97.4} \\
\cmidrule{1-5}

\multirow{3}{*}{\textbf{DyCAST-BP1}}
& {Tokens + durations} & \textbf{3.94} & \textbf{2.37} & \textbf{97.7} \\
& {Tokens + utterance length} & 3.92 & \underline{3.61} & 97.6 \\
& {Tokens only} & \underline{3.93} & 4.49 & 97.6 \\
\cmidrule{1-5}

\multirow{3}{*}{\textbf{DyCAST-BP3}}
& {Tokens + durations} & \underline{3.95} & \textbf{2.99} & \textbf{97.4} \\
& {Tokens + utterance length} & \underline{3.95} & \underline{4.66} & \underline{97.2} \\
& {Tokens only} & \textbf{3.96} & 5.13 & \underline{97.2} \\
\cmidrule{1-5}

\multirow{3}{*}{\textbf{DyCAST-BP5}}
& {Tokens + durations} & 3.93 & \textbf{5.12} & \textbf{96.6} \\
& {Tokens + utterance length} & \underline{3.97} & 8.84 & \underline{96.5} \\
& {Tokens only} & \textbf{3.98} & \underline{8.67} & \underline{96.5} \\

\bottomrule
\end{tabular}
}
\vskip -0.05in
\end{table*}

\subsection{Qualitative Analysis of DyCAST Chunk Boundaries}
\label{sec:appendix_spectrograms}

\begin{figure*}[h]
\centering

\resizebox{\textwidth}{!}{%
\begin{minipage}{\textwidth}

\makebox[0.24\textwidth][c]{\textbf{DyCAST-CA}} \hfill
\makebox[0.24\textwidth][c]{\textbf{DyCAST-BP1}} \hfill
\makebox[0.24\textwidth][c]{\textbf{DyCAST-BP3}} \hfill
\makebox[0.24\textwidth][c]{\textbf{DyCAST-BP5}}

\vspace{0.1em}

\begin{subfigure}[t]{0.24\textwidth}\centering
\includegraphics[width=\linewidth]{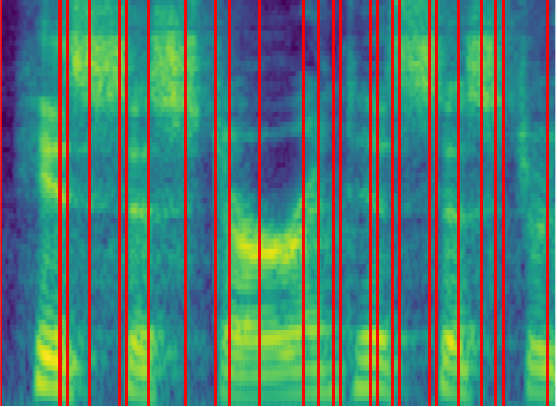}
\end{subfigure}\hfill
\begin{subfigure}[t]{0.24\textwidth}\centering
\includegraphics[width=\linewidth]{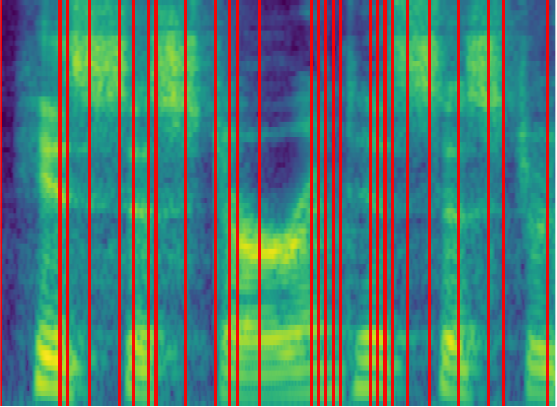}
\end{subfigure}\hfill
\begin{subfigure}[t]{0.24\textwidth}\centering
\includegraphics[width=\linewidth]{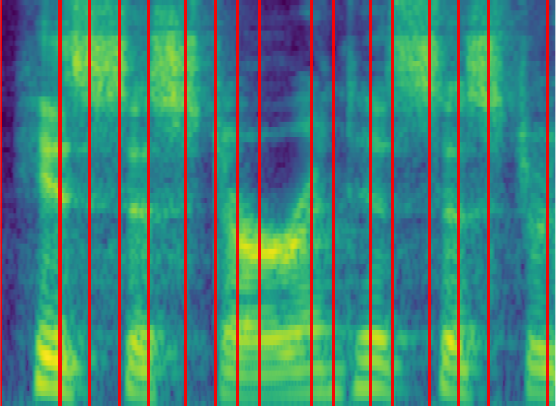}
\end{subfigure}\hfill
\begin{subfigure}[t]{0.24\textwidth}\centering
\includegraphics[width=\linewidth]{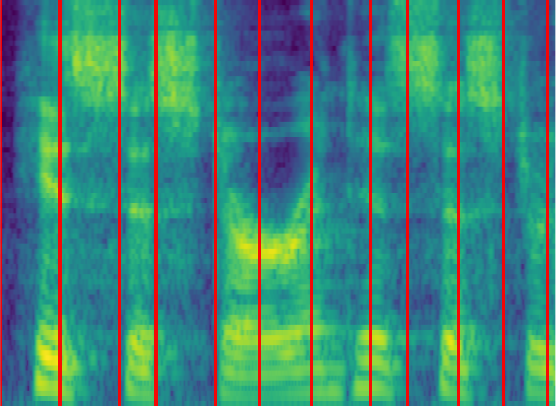}
\end{subfigure}

\end{minipage}%
} 

\vspace{-0.1cm}
\caption{DyCAST chunk boundaries at average frame rates of approximately 14 Hz, 17 Hz, 9 Hz, and 6 Hz (from left to right).}
\label{fig:spectrograms}
\vspace{-0.5cm}
\end{figure*}

\cref{fig:spectrograms} provides qualitative spectrogram visualizations of DyCAST chunk boundaries for the four variants considered in this work: {DyCAST-CA}, {DyCAST-BP1}, {DyCAST-BP3}, and {DyCAST-BP5}. DyCAST-CA relies on boundaries provided by the character aligner, while the remaining variants use the boundary predictor with \texttt{min\_gap} set to 1, 3, and 5, respectively, resulting in average frame rates of approximately 14 Hz, 17 Hz, 9 Hz, and 6 Hz. As the frame rate decreases, chunk boundaries become less frequent and more widely spaced, reflecting longer token durations and coarser temporal resolution.

\end{document}